# NEURAL SURROGATE FORWARD MODELLING FOR ELECTROCARDIOLOGY WITHOUT EXPLICIT INTRACELLUAR CONDUCTIVITY TENSOR

**Shaheim Ogbomo-Harmitt[1], Cesare Magnetti[2], Jakub Grzelak[1], and Oleg Aslanidi[1]**
[1] King's College London, Floor 3, Lambeth Wing St Thomas' Hospital Westminster Bridge Road, London, SE1 7EH, {shaheim.ogbomo-harmitt, jakub.grzelak, oleg.aslanidi}@kcl.ac.uk
[2] PhysicsX, 1 Leonard Circus, London EC2A 4DQ, cesare.magnetti@physicsx.ai

**SUMMARY**

Accurate forward modelling is essential for non-invasive cardiac electrophysiology, particularly in atrial fibrillation, where electrical activation is highly disorganised. Conventional physics-based forward models require explicit specification of intracellular conductivity tensors, which are not directly measurable in clinical practice and introduce structural modelling errors. This proof-of-concept study presents a deep learning approach that learns a direct mapping from left atrial intracellular electrical potentials to far-field ECGs without requiring explicit intracellular conductivity inputs at inference time. Despite training only on 74 subjects, the model achieved an $R^2$ of $0.949 \pm 0.037$, highlighting potential to reduce structural uncertainty and improve non-invasive AF assessment.

**Key words:** *Electrocardiology, Forward Solution, Deep Learning, Atrial Fibrillation*

## 1 INTRODUCTION

The forward problem in electrocardiology is the task of predicting body surface potentials from a given distribution of cardiac electrical activity. Its solution underpins the inverse problem, which aims to reconstruct cardiac activation from body surface potentials. Accurate forward modelling is therefore essential for non-invasive evaluation of cardiac electrical activity, particularly in atrial fibrillation (AF), where electrical activation is highly disorganised. Non-invasive assessment of atrial electrophysiology has the potential to guide catheter ablation and thereby improve outcomes in persistent AF, which continues to show suboptimal treatment success. Current forward solutions use physics-based models of intracellular electrical propagation and torso conduction. These models require specification of the intracellular conductivity tensor, whose components include tissue conductivity, myocardial fibre direction and anisotropy ratio. However, these properties cannot be measured in routine clinical practice and must be assumed or, at best, estimated. This introduces structural modelling errors that propagate directly into the inverse solution and limit the accuracy of non-invasive reconstructions. Previous work has mainly focused on the integration of inhomogeneous torso electrical properties in the forward model to reduce these discrepancies, but this has only achieved modest gains [1]. An alternative strategy is to address the error introduced by the assumed intracellular conductivity tensor. This proof-of-concept study investigates whether a deep learning (DL) model can bypass this source of error by learning a direct mapping from intracellular atrial activation to far field extracellular electrocardiogram (ECG) morphology. Because the learned forward surrogate does not require explicit specification of the intracellular conductivity tensor at inference, it has the potential to reduce cardiac tissue assumption error. Reducing this structural modelling uncertainty may improve the accuracy of inverse atrial electrophysiology solutions, particularly in the challenging context of AF.

## 2 METHODOLOGY

Atrial electrophysiology was simulated in openCARP using a monodomain model with the AF-adapted Courtemanche–Ramírez–Nattel cell formulation. Conduction velocity was set to 0.7 $ms^{-1}$ with a 4:1 longitudinal–transverse anisotropy. 93 patient-specific left atrial (LA) surface meshes from Roney et al. were used (seven excluded for instability from original dataset), with fibre orientations mapped from an atlas [2]. Thus, spatially varying and subject-specific conductivity tensor fields were generated by the geometry-dependent fibre orientations, despite fixed conductivity magnitudes and anisotropy ratios across simulations. Lead II ECGs were computed via the infinite-volume conductor method assuming a homogeneous conductivity of 0.24 $S \cdot m^{-1}$ and scaled by a 1.4 mm surface-to-volume factor for monolayer representation. Lead II was derived using the Grzelak et al. torso model with anatomically registered electrode positions, and signals were band-pass filtered between 0.05 and 60 Hz using second-order Bessel filters, consistent with clinical ECG filtering [3]. The proposed DL framework employs a sequence-to-sequence architecture: comprising an encoder and a decoder to translate a series of voltage propagation maps into a Lead II ECG signal. The encoder maps the 3D coordinates and transmembrane voltages defined on the atrial surface to a fixed-dimensional latent vector. Each mesh node provides voltage $V_i \in \mathbb{R}$. These inputs are transformed through three feature streams: local geometric encoding, global spectral propagation, raw voltage projection, feature stream gating and quadrature (Figure. 1).

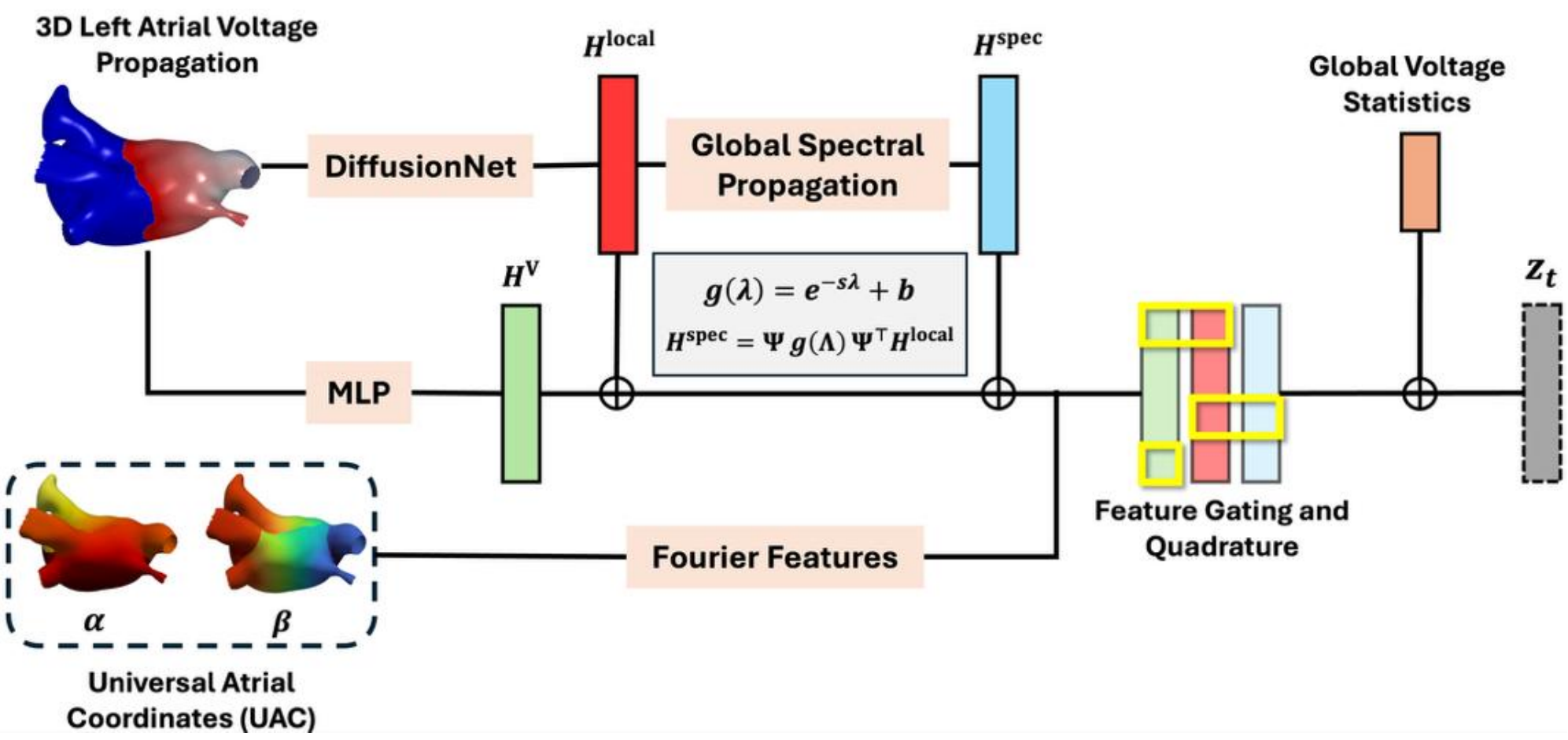

**Figure 1:** Encoder architecture that maps atrial-surface voltages to a latent vector by combining three different feature streams, fused by learned gating and summarised through mass-weighted pooling.

Local geometry-aware information is extracted using DiffusionNet, which is robust to changes in mesh density and resolution. The method relies on precomputed geometric operators: the cotangent Laplacian $L$, mass matrix $M$, intrinsic gradient operators $\nabla$ and the Laplace–Beltrami eigenvalues and eigenfunctions $(\lambda, \phi)$ [4]. Given these fixed operators, each node is mapped to a local embedding $h_i^{\text{local}} = F_{\text{DiffusionNet}}(V_i, L, M, \nabla, \lambda, \phi)$. To complement the local features with non-local geometric structure, we apply a global spectral propagation to the DiffusionNet embeddings ($H^{\text{local}}$). If we let $\Psi = [\psi_1, \dots, \psi_K]$ denote the first $K = 64$ Laplace–Beltrami eigenfunctions on the atrial surface, and $\Lambda = \text{diag}(\lambda_1, \dots, \lambda_K)$ the corresponding diagonal matrix of eigenvalues. These eigenpairs form a smooth harmonic basis over the atrial surface, where low frequency modes capture broad geometric structure and higher frequencies encode finer detail. Using this basis, the global spectral representation of the local features $H^{\text{local}}$ is computed as in Equation. 1, where a learnable scalar response $g(\Lambda)$ is applied to each eigenvalue. The spectral response $g(\lambda)$ is parametrised as in Equation. 2, with parameters $s$ and $b$ learned during training to control the attenuation or amplification of different Laplace–Beltrami frequencies. The bias term $b$ adds a non-decaying global spectral component by uniformly weighting the retained Laplace–Beltrami eigenmodes. With $b > 0$ (here $b = 1,\ s = 0.1$ in trained model), it bypasses heat-kernel attenuation, giving an explicit global projection over the atrial surface. This provides efficient long-range, geometry-aware coupling in one step, unlike stacked diffusion blocks that spread information gradually and can oversmooth.

$$H^{\text{spec}} = \Psi\, g(\Lambda)\, \Psi^{\top} H^{\text{local}} \tag{1}$$

$$g(\lambda) = e^{-s\lambda} + b \tag{2}$$

Moreover, a lightweight multilayer perceptron (MLP) maps each voltage value to a feature vector, $h_i^V = \text{MLP}(V_i)$, ensuring that sharp voltage transitions remain represented even if geometric processing induces smoothing. The three feature streams are then combined using a per-node gating mechanism. For each node, a small neural network outputs three softmax-normalised coefficients $(g_i^{\text{local}}, g_i^{\text{spec}}, g_i^V)$, which determine the relative contribution of the local, spectral, and voltage-derived features. The fused feature at each $i_{th}$ node is then computed as $h_i^{\text{fused}} = g_i^{\text{local}} h_i^{\text{local}} + g_i^{\text{spec}} h_i^{\text{spec}} + g_i^V h_i^V$. To summarise the fused node features into a single latent vector, the encoder applies a quadrature procedure guided by the Universal Atrial Coordinate (UAC) system [5]. Each node's UAC coordinates, $(\alpha_i, \beta_i)$, are first encoded using multi-frequency Fourier features $(\gamma_i)$, which capture smooth spatial variations across the LA surface, with $L = 4$ in our implementation (Equation. 3).

$$\gamma_i(\alpha_i, \beta_i) = \begin{bmatrix} \phi_0(\alpha_i, \beta_i) \\ \phi_1(\alpha_i, \beta_i) \\ \vdots \\ \phi_{L-1}(\alpha_i, \beta_i) \end{bmatrix}, \ \phi_k(\alpha_i, \beta_i) = \begin{bmatrix} \sin(2^k \pi \alpha_i) \\ \cos(2^k \pi \alpha_i) \\ \sin(2^k \pi \beta_i) \\ \cos(2^k \pi \beta_i) \end{bmatrix}, \ k = 0, \cdots, L-1 \tag{3}$$

Fourier-encoded UAC features $(\alpha_i, \beta_i)$ are passed through a network that assigns each node a scalar importance score, which is then normalised by the nodal masses so that contributions reflect true geometric area rather than sampling density. These mass-normalised weights determine each node's influence in the geometry-aware pooled features, formed as a weighted combination of the fused node embeddings. To complement this pooled representation, the encoder also computes global voltage statistics (median, median absolute deviation, mass-weighted mean and standard deviation). These statistics are concatenated with the pooled feature and passed through a final projection layer to produce the latent vector. The DL model's decoder maps the latent sequence $Z = \{z_t\}_{t=1}^T$ to a lead II ECG trace using sinusoidal time embeddings, time-aware Bahdanau attention and a Long Short-Term Memory (LSTM) network (Figure. 2). Each sinusoidal embedding $e_t$ provides a smooth, multi-scale representation of temporal position, enabling the model to encode both short- and long-range timing structure.

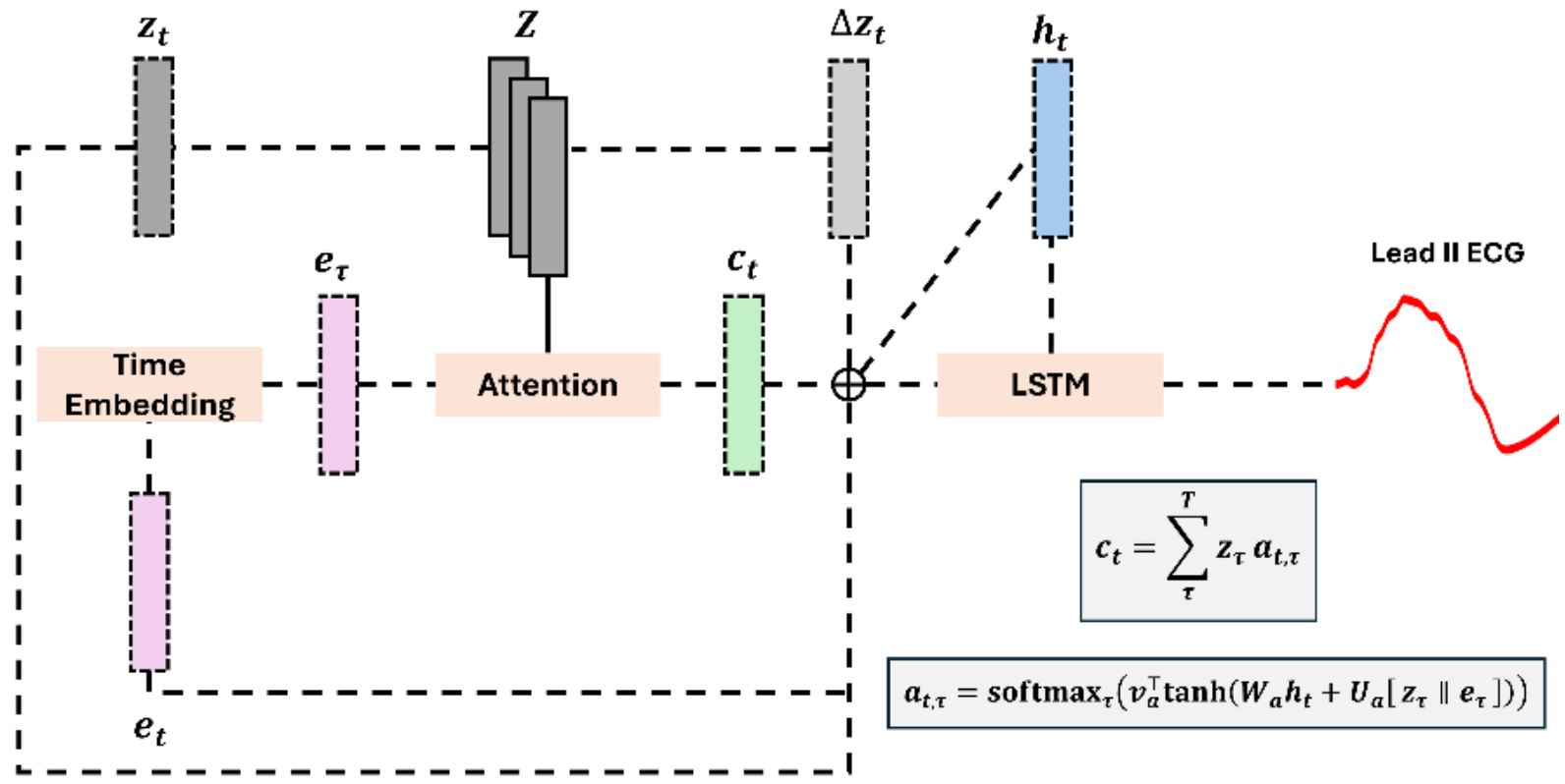

**Figure 2:** Decoder architecture that predicts the Lead II ECG trace from latent sequences using sinusoidal time embeddings, time-aware attention, latent-difference augmentation and LSTM.

At decoding step $t$, the decoder hidden state $h_t$ attends over all latent vectors in $Z$. Since attention requires comparing the decoder state at time $t$ with every encoded time step, we introduce the index $\tau$. The attention weights are computed according to Equation. 4, where $W_a$, $U_a$, and $v_a$ are learned parameters and $\parallel$ denotes feature concatenation.

$$a_{t,\tau} = softmax_\tau\left(v_a^\top tanh(W_a h_t + U_a[z_\tau \parallel e_\tau])\right) \tag{4}$$

These weights define how strongly the decoder attends to each latent vector when predicting the next ECG value. Using them, the model forms a context vector $c_t$, which is a weighted combination of all latent representations and, therefore, reflects the temporal regions of the sequence that are most informative for step $t$. To capture rapid temporal transitions in the latent space, the decoder further computes latent differences $\Delta z_t = z_t - z_{t-1}$. At each time step, the inputs $(c_t, z_t, \Delta z_t, e_t)$ are concatenated and passed to a LSTM network. The LSTM output, together with the previous ECG prediction, is processed through fully connected layers to generate the next ECG value. This autoregressive procedure is repeated for all $t = 1, \ldots, T$ to produce the complete ECG waveform. Data were split into 80% training, 10% validation and 10% testing. ECGs and atrial voltages were z-score normalised. The model was trained for 50 epochs using Adam (initial learning rate 0.001, halved every three epochs), and the checkpoint with the highest validation $R^2$ was selected. Implementation used PyTorch with NumPy. Training employed a hybrid loss combining a Huber time-domain loss with a spectral-entropy loss that penalises differences in the Shannon entropy of normalised power spectral densities. The total loss was $L_{\text{total}} = L_H + \omega(n)\, L_{\text{SE}}$, where $\omega(n)$ follows a cosine decay to progressively down-weight spectral regularisation [6]. An ablation study was carried out and showed that removing components of encoder lead to decrease in performance.

## 3 RESULTS AND CONCLUSIONS

As shown in Figure 3, the DL model achieved an $R^2$score of 0.949 ± 0.037 and a mean absolute error of 0.0036 ± 0.0015 mV on the test set, demonstrating the effectiveness of the proposed framework for predicting extracellular potentials without requiring explicit specification of intracellular conductivity tensors as model inputs. Unlike conventional physics-based forward solvers, which necessitate explicit conductivity parameterisation, the proposed approach learns an effective forward mapping directly from data. Notably, this level of predictive accuracy was achieved despite the model being trained on data from only 74 subjects. This suggests that incorporating a substantially larger and more diverse training cohort could further improve predictive performance and generalisability. Importantly, although the training data were generated using monodomain simulations with fixed conduction velocity and anisotropy ratios, subject-specific fibre orientations resulted in spatially varying and geometry-dependent conductivity tensor fields across the cohort. The DL model therefore learns to account for anisotropic conduction effects implicitly, without requiring explicit conductivity modelling at inference time. Thus, this study provides a proof-of-concept for a data-driven forward solution in electrocardiology that avoids explicit intracellular conductivity specification, with potential implications for improving ECG imaging accuracy and the non-invasive identification of AF ablation targets. Future work will focus on extending the framework to whole-heart or biatrial geometries (particularly for AF applications), predicting multiple ECG leads simultaneously and modelling ECGs associated with reentrant electrical activity.

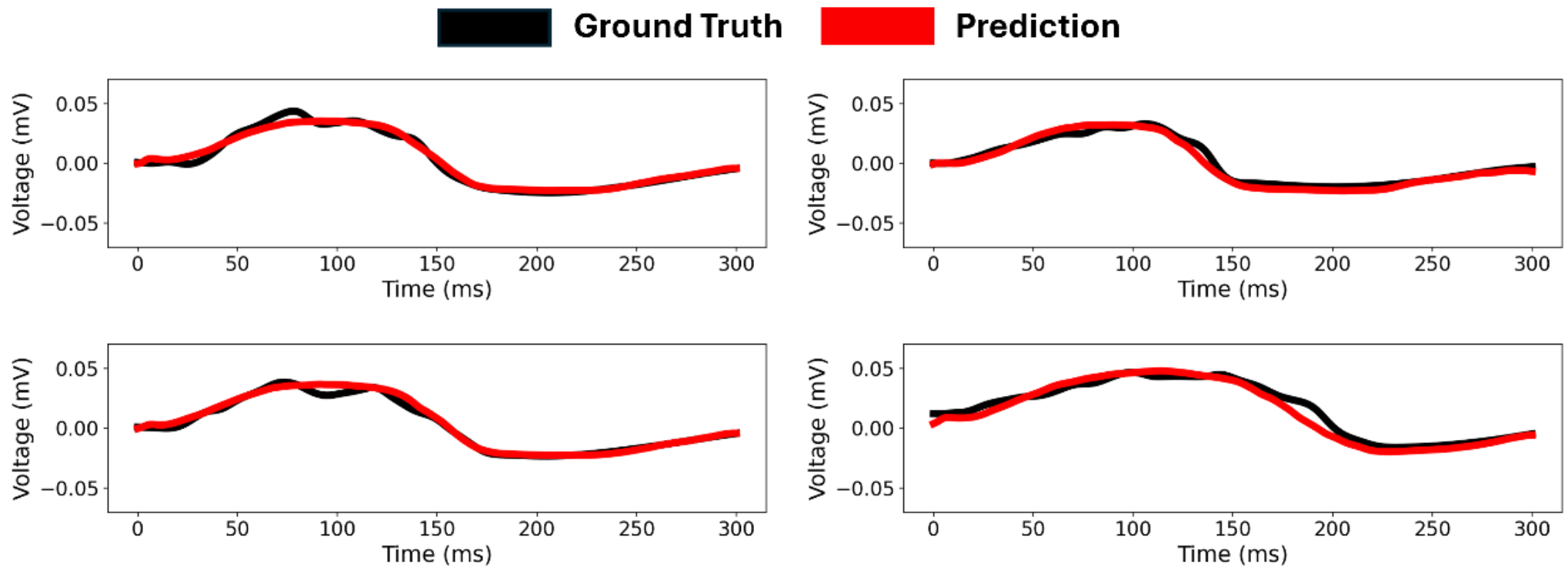

**Figure 3.** Ground-truth (black) and deep learning–predicted (red) left atrial Lead II ECG signals for representative examples, showing close agreement in waveform morphology and timing.